\documentclass[11pt]{article}
\usepackage{amsmath}
\usepackage{amssymb}
\usepackage{algorithm}
\usepackage{algorithmic}
\usepackage{graphicx}
\usepackage{subcaption}
\usepackage{hyperref}
\usepackage{booktabs}
\usepackage{multirow}
\usepackage{array}
\usepackage{url}
\usepackage{cite} 

\title{Deep Learning-Based Burned Area Mapping Using Bi-Temporal Siamese Networks and AlphaEarth Foundation Datasets}

\author{Seyd Teymoor Seydi \\ School of Surveying and Geospatial Engineering, College of Engineering, \\ University of Tehran, Tehran 14174-66191, Iran\\ seydi.teymoor@ut.ac.ir}
\begin{document}

\maketitle
\begin{abstract}

Accurate and timely mapping of burned areas is crucial for environmental monitoring, disaster management, and assessment of climate change. This study presents a novel approach to automated burned area mapping using the AlphaEArth dataset combined with the Siamese U-Net deep learning architecture. The AlphaEArth Dataset, comprising high-resolution optical and thermal infrared imagery with comprehensive ground-truth annotations, provides an unprecedented resource for training robust burned area detection models. We trained our model with the Monitoring Trends in Burn Severity (MTBS) dataset in the contiguous US and evaluated it with 17 regions cross in Europe.  Our experimental results demonstrate that the proposed ensemble approach achieves superior performance with an overall accuracy of 95\%, IoU of 0.6, and F1-score of 74\% on the test dataset. The model successfully identifies burned areas across diverse ecosystems with complex background, showing particular strength in detecting partially burned vegetation and fire boundaries and its transferability and high generalization in burned area mapping. This research contributes to the advancement of automated fire damage assessment and provides a scalable solution for global burn area monitoring using the AlphaEarth dataset.

\end{abstract}
\section{Introduction}

The mapping of burned areas from satellite imagery has become increasingly critical as wildfire frequency and severity continue to escalate worldwide \cite{seydi2025dlsr}. Conventional change detection methodologies employing standard multispectral data typically demonstrate reasonable efficacy under optimal conditions \cite{filipponi2018bais2, martin2005performance}. However, these methodologies frequently encounter challenges in accurately discerning the subtleties present in real-world optical imagery \cite{giglio2009active}. Factors such as cloud cover, smoke plumes, atmospheric scattering, seasonal vegetation changes, and terrain-induced shadows introduce significant noise and spectral ambiguity, leading to missed detections and false alarms \cite{roy2008collection}. These issues have the effect of reducing the reliability and applicability of burned area detection, particularly in regions characterized by persistent cloudiness or heterogeneous landscapes.

To address these challenges, this work leverages the AlphaEarth Foundations dataset, a cutting-edge advancement in Earth observation that provides embedding field models synthesizing spatial, temporal, and contextual information from multiple sources \cite{brown2025alphaearth}. Rather than relying solely on raw spectral measurements, AlphaEarth embeddings encode rich geospatial patterns that are resilient to noise, seasonal variability, and partial occlusions caused by clouds or atmospheric artifacts \cite{houriez2025scalablegeospatialdatageneration}. This approach enables a more comprehensive representation of land surface dynamics, ranging from local burned patches to continental-scale monitoring \cite{murakami2025within}. A notable strength of AlphaEarth is its capacity to function effectively even in scenarios where labeled training data is limited, a prevalent challenge in remote sensing applications. The dataset covers the period from 2017 to 2024, providing embedding field layers ready for annual analysis that encapsulate the comprehensive complexity of the Earth's surface and its temporal evolution \cite{brown2025alphaearth}. Each year comprises 64 bands of information, thereby encompassing intricate environmental interactions across multiple modalities. The dataset documentation does not explicitly map each band to a specific month, which presents an additional layer of complexity requiring careful preprocessing and implementation strategies \cite{brown2025alphaearth}. The integration of this sophisticated geospatial embedding with advanced learning models is intended to enhance the accuracy, generalization, and scalability of burned area mapping, even under challenging optical conditions \cite{murakami2025within}.

The AlphaEarth dataset offers high-dimensional feature representations, including optical and thermal imagery. Although these representations are rich in information, they can result in computationally expensive models and an elevated risk of overfitting, particularly when the availability of labeled samples is limited. In order to address these challenges, a lightweight U-Net backbone was designed. This backbone reduces the number of trainable parameters while maintaining strong feature extraction capabilities. In the context of this research, a Siamese network architecture was implemented to detect burned areas. This architecture performs multi-scale differencing of feature maps. The Siamese approach, in contrast to conventional pixel-based comparison methods, enables the model to discern the intricate spatial and spectral transformations induced by fires in vegetation, soil, and other components of the landscape. By capturing both local and contextual changes across multiple scales, the network can robustly identify burned regions, even in heterogeneous landscapes. The integration of the lightweight U-Net with the Siamese multi-scale differencing strategy ensures the accurate, generalizable, and computationally efficient detection of burned area. This approach is capable of handling high-dimensional data while maintaining sensitivity to subtle fire-induced changes and complex fire boundaries.

The real test of our approach comes from its cross-continental deployment strategy. We trained exclusively on fires from the central United States using the Monitoring Trends in Burn Severity (MTBS) dataset \cite{eidenshink2007project}, then threw the model into completely unseen territory: European fire events documented by the Copernicus Emergency Management Service (EMS) \cite{copernicus}. This geographic leap tests whether the model truly learns universal fire signatures or merely memorizes regional patterns—a critical distinction for any operational system.

\section{Dataset and Study Area}

\subsection{Dataset}
The AlphaEarth Foundations dataset signifies a paradigm shift in the utilization of Earth observation data. AlphaEarth does not provide raw spectral measurements; rather, it delivers embedding field models. These models are pre-processed representations that integrate multiple satellite observations, atmospheric corrections, and contextual features. This design significantly reduces the necessity for manual data fusion, thereby enabling researchers to prioritize task-specific modeling rather than the preliminary processing of data. The dataset extends from 2017 to 2024, providing annual composites that encapsulate the dynamics of surface features over time. Each annual layer comprises 64 embedding bands. While these bands encapsulate seasonal and temporal variation, the documentation does not explicitly assign bands to particular months or time intervals. This ambiguity poses challenges for applications such as fire disturbance mapping, where intra-annual timing is crucial. However, it also encourages the development of models capable of learning temporal dynamics directly from embedding features. To this end, we used bitemporal datasets (pre/post-fire). The technical details of the dataset utilized in this study are delineated in Table~\ref{tab:dataset_info}.

\begin{table}[h!]
\centering
\caption{Technical specifications of the AlphaEarth Foundations dataset.}
\label{tab:dataset_info}
\resizebox{\textwidth}{!}{%
\begin{tabular}{l l}
\hline
\textbf{Property} & \textbf{Description} \\
\hline
Source & AlphaEarth Foundations (Google DeepMind) \\
Coverage Years & 2017–2024 \\
Spatial Resolution & 10 m (embedding field representation) \\
Spectral/Embedding Bands & 64 embedding bands per year \\
Temporal Coverage & Annual composites (multi-seasonal embeddings) \\
Preprocessing & Multi-sensor fusion, atmospheric correction, contextual integration \\
Data Type & Embedding representations (not raw reflectance) \\

\hline
\end{tabular}%
}
\end{table}

\subsection{Study Area}
The objective of the present study was to assess the generalizability of the proposed model across distinct geographical locations. The training and validation phase was conducted using the Monitoring Trends in Burn Severity (MTBS) dataset \cite{eidenshink2007project}, which provides historical fire perimeters and severity assessments across the central United States. The region under study is characterized by a combination of grassland, agricultural, and forested ecosystems, which collectively provide a diverse array of fire regimes conducive to the robust calibration of models. To ensure the model's independence, a cross-continental generalization test was conducted (17 Sites). The model was evaluated in regions across Europe, thereby providing an independent evaluation of its performance (Table~\ref{tab:study_area}). The European test areas exhibit significant ecological distinctions from the U.S. training domain, rendering them suitable for the evaluation of the model's transferability and robustness.

\begin{table}[h!]
\centering
\caption{European EMSR test sites used for model evaluation}
\label{tab:study_area}
\resizebox{\textwidth}{!}{%
\begin{tabular}{l l l l}
\toprule
\textbf{Site} & \textbf{Country} & \textbf{Location} & \textbf{Date of Fire} \\
\midrule
EMSR633 & France & Saumos, Gironde & 12-14 Sep 2022 \\
EMSR749 & Greece & Kavala, East Macedonia & 22-24 Aug 2024 \\
EMSR748 & Portugal & Central Madeira & 14-18 Aug 2024 \\
EMSR747 & Greece & Central Macedonia & 9-14 Aug 2024 \\
EMSR746 & Greece & Attica & 11-12 Aug 2024 \\
EMSR745 & Greece & Crete & 7-8 Aug 2024 \\
EMSR744 & Greece & Evia Island & 31 Jul-01 Aug 2024 \\
EMSR740 & Bulgaria & Slavyanka Mountain & 24 Jul 2024 \\
EMSR738 & Bulgaria & Yambol Region & 18 Jul 2024 \\
EMSR737 & Greece & Korinthia & 11-13 Jul 2024 \\
EMSR736 & Albania & Gjirokastra and Vlora & 9-10 Jul 2024 \\
EMSR735 & Greece & Kos Island & 1-2 Jul 2024 \\
EMSR733 & Greece & Attica & 29 Jun-01 Jul 2024 \\
EMSR731 & Greece & Alpochori & 21-23 Jun 2024 \\
EMSR730 & Greece & Latas & 21-22 Jun 2024 \\
EMSR729 & Greece & Kotopi & 19 Jun 2024 \\
EMSR638 & Greece & Papikio Mountain & 22 Oct-03 Nov 2022 \\
\bottomrule
\end{tabular}%
}
\end{table}

\begin{figure}[p]
    \centering
    \makebox[\textwidth][c]{\includegraphics[width=1.2\textwidth]{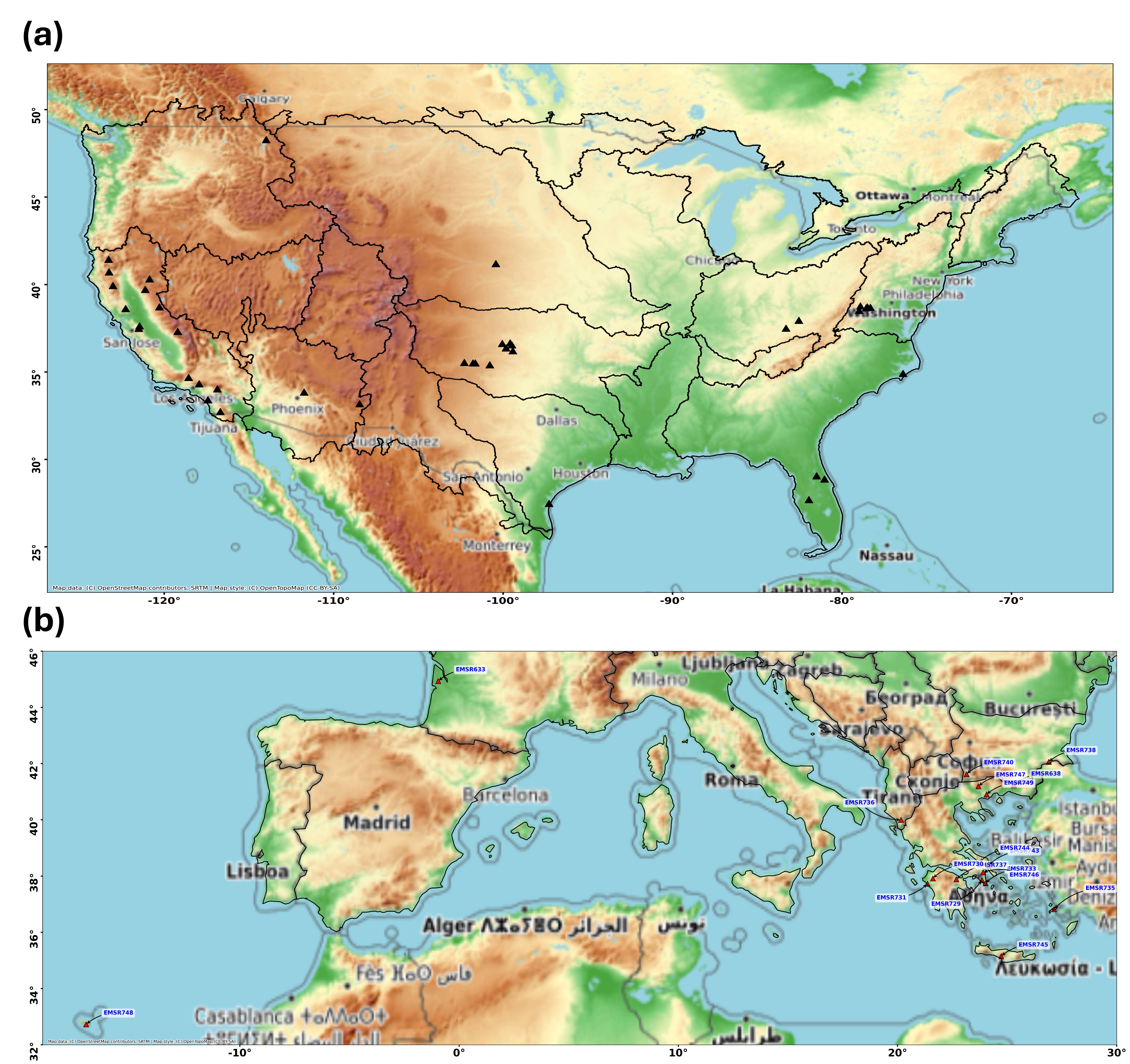}}
    \caption{Geographical distribution of the datasets used in this study. 
    Panel (a) shows the central United States, where the MTBS dataset was used for training and validation, 
    while panel (b) highlights the European regions used for independent testing.}
    \label{fig:dataset_locations}
\end{figure}

\section{Methodology}

We developed a Siamese U-Net architecture for the detection of burned areas from bi-temporal satellite imagery. This architecture utilizes U-Net's encoder-decoder structure to preserve spatial context, while employing weight-sharing between parallel branches to process pre- and post-fire imagery~\cite{ronneberger2015u}. The network processes paired inputs, specifically pre-fire imagery at time $t_1$ and post-fire imagery at time $t_2$. It has been demonstrated that both temporal inputs pass through identical encoder branches with shared convolutional weights, ensuring consistent feature extraction across time points. This weight-sharing mechanism enables the network to learn representations that are invariant to temporal changes while remaining sensitive to fire-induced alterations.

The encoder pathway consists of repeated application of convolutional blocks, where each block $E_l$ at level $l \in \{1, 2, 3, 4, 5\}$ performs:
\begin{equation}
E_l(x) = \text{MaxPool}(\text{ReLU}(\text{BN}(\text{Conv}_{3\times3}(x))))
\end{equation}
where $\text{Conv}_{3\times3}$ denotes a $3\times3$ convolution, $\text{BN}$ represents batch normalization, and $\text{MaxPool}$ performs $2\times2$ max pooling with stride 2.

The decoder pathway implements explicit change detection through feature differencing. At each decoder level $l$, the network computes element-wise subtraction between encoded features:
\begin{equation}
\Delta F_l = F_{\text{post}}^l - F_{\text{pre}}^l
\end{equation}

The deepest decoder layer receives the difference features from the bottleneck layer, applies transposed convolution for spatial upsampling, and propagates to subsequent layers. Each decoder block $D_l$ performs:
\begin{equation}
D_l = \text{Conv}_{3\times3}(\text{Concat}[\text{UpConv}_{2\times2}(\Delta F_{l+1}), \Delta F_l^{\text{skip}}])
\end{equation}
where $\text{UpConv}_{2\times2}$ denotes transposed convolution for upsampling and $\Delta F_l^{\text{skip}}$ represents the skip-connected difference features from the corresponding encoder level.

This multi-scale differencing strategy, combined with skip connections at resolutions $\{1/2, 1/4, 1/8, 1/16\}$ of the original input, enables precise delineation of burn boundaries across heterogeneous landscapes.

To address the severe class imbalance inherent in burned area mapping, we employ a hybrid loss function combining weighted binary cross-entropy (BCE) and Dice coefficient:
\begin{equation}
\mathcal{L}_{\text{total}} = \alpha \cdot \mathcal{L}_{\text{BCE}} + (1-\alpha) \cdot \mathcal{L}_{\text{Dice}}
\end{equation}

\section{Methodology}

We created a special type of deep learning network called a U-Net-based Siamese network that can detect wildfires from satellite images. The network combines the strengths of U-Net for capturing spatial context with a Siamese design that explicitly compares images before and after a fire . The model takes two types of inputs: images captured before and after the fire. It uses these inputs to extract feature maps through a shared-weight convolutional backbone \cite{ronneberger2015u}. This lets the network focus on the changes caused by the fire while keeping the same representation of the land. In the decoder, the reconstruction process highlights the differences between feature maps. The first decoder layer receives the different features from the last encoder layer, upsamples them, and passes them to the next decoder layer. In the next layers, including the second decoder layer, the model combines the output from the previous decoder layer with skip-connected feature maps from the encoder. It does this to find the differences and highlight burn-affected regions. This combination of differences at different sizes and connections between them preserves fine spatial details while emphasizing fire-induced changes, enabling accurate and robust mapping across different landscapes and environmental conditions.

In addressing the data imbalance that is intrinsic to the process of burned area mapping, a methodology has been employed that utilizes a weighted combination of binary cross-entropy and Dice losses. The following definition can be posited for this methodology:

\begin{equation}
\mathcal{L}_{total} = \mathcal{L}_{BCE} + \lambda_{dice} \cdot \mathcal{L}_{Dice}
\end{equation}

\section{Experimental Results}

To illustrate the resilience of the proposed Siamese U-Net architecture for burned area mapping, three notable European EMSR test sites were selected from a validation set of 17 sites: The following three EMSR codes have been identified: EMSR744, EMSR746, and EMSR733. As illustrated in Figure~\ref{fig:burned_area_visualization}, the model's predictive performance is demonstrated through the use of confusion matrix visualizations, which reveal distinct patterns in classification accuracy. The model demonstrates notable proficiency in delineating edges, accurately depicting fire perimeter boundaries with minimal edge bleeding artifacts. True positive (TP) regions demonstrate robust detection of core burned areas, while false negatives (FN) predominantly occur along transition zones where burn severity gradients create spectral ambiguity. These misidentifications are particularly evident in EMSR746's northwestern regions, where partial canopy damage and understory burns challenge the binary classification threshold. False positives (FP) manifest primarily in three scenarios: shadow-induced misclassifications in mountainous terrain (visible in EMSR733), phenological changes in agricultural fields mimicking burn signatures, and bare soil patches exhibiting similar spectral characteristics to ash-covered surfaces. The model demonstrates superior performance in homogeneous burn scars; however, it encounters challenges in heterogeneous landscapes comprising mixed severity burns. Edge accuracy analysis reveals sub-pixel precision in high-contrast boundaries but degraded performance in gradual burn transitions. EMSR744 demonstrates optimal detection with minimal commission errors, while EMSR746 exhibits increased omission errors in fragmented burn patterns. This observation underscores the trade-off between sensitivity and specificity in operational burned area mapping.

\begin{figure}[p]
    \centering
    \makebox[\textwidth][c]{\includegraphics[width=1.1\textwidth]{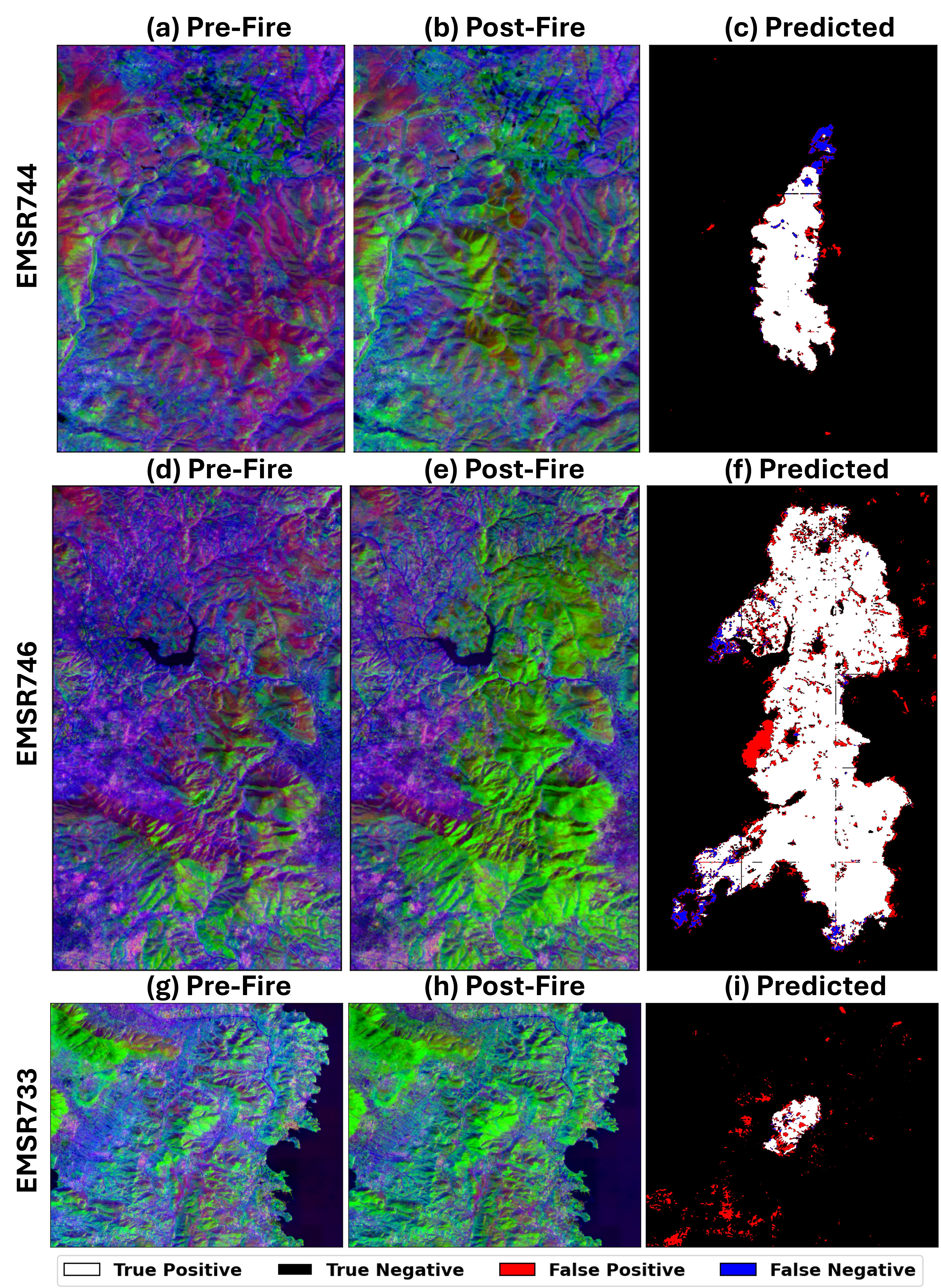}}
    \caption{Visual results of burned area detection for three representative European EMSR sites. 
    Columns correspond to (from left to right) pre-fire imagery, post-fire imagery, and the predicted burned area maps from the Siamese U-Net model. 
    Rows correspond to different test sites: EMSR744 (row 1), EMSR746 (row 2), and EMSR733 (row 3). 
    The model accurately delineates burned regions in EMSR744 and EMSR746, capturing complex fire boundaries and partially burned vegetation.}
    \label{fig:burned_area_visualization}
\end{figure}

The model demonstrates robust performance across the 17 European EMSR test sites, achieving a mean overall accuracy of 0.949 ± 0.052 (see Table~ref{tab:results}). The comprehensive evaluation reveals nuanced performance characteristics: precision of 0.806±0.211, recall of 0.749±0.231, F1-score of 0.735±0.186, IoU of 0.612±0.220, and Dice coefficient of 0.735±0.186. The high overall accuracy, coupled with moderate overlap metrics, indicates a class imbalance phenomenon typical in burned area mapping, where unburned pixels dominate the landscape, artificially inflating pixel-wise accuracy. The more stringent intersection over union (IoU) metric reveals challenges in precise burned area delineation. The substantial variability in precision and recall across sites indicates the model's sensitivity to fire regime characteristics and landscape heterogeneity. This performance heterogeneity manifests through distinct failure modes: commission errors (low precision) predominantly occur in agricultural landscapes and urban-wildland interfaces where spectral confusion with harvested fields, bare soil, and shadows creates false positives. Omission errors, characterized by low recall, have been observed to concentrate in areas exhibiting signs of partial canopy mortality, understory burns, and rapid post-fire greening. In such environments, the spectral signal attenuation poses a significant challenge to detection thresholds. A site-specific analysis reveals compelling patterns in model behavior. EMSR731 (Alpochori, Greece) demonstrates near-perfect pixel accuracy (0.997) but paradoxically low recall (0.417), indicating that the model conservatively underestimates burn extent while maintaining high confidence in detected areas. This characteristic is indicative of well-defined, high-severity burns with sharp boundaries. In contrast, EMSR747 (Central Macedonia) demonstrates catastrophic precision collapse (0.353) despite high recall (0.974), suggesting aggressive over-detection likely initiated by complex topographic shadows and mixed vegetation communities generating burn-like spectral signatures.
The precision-recall trade-off manifests pronounced geographical clustering. Mediterranean sites characterized by homogeneous shrubland burns (EMSR744: F1=0.932, IoU=0.873) demonstrate optimal balance, while mountainous regions exhibiting fragmented forest fires exhibit degraded performance. EMSR733 exemplifies this challenge with extreme precision-recall imbalance (0.456 vs. 0.940), where topographic effects and heterogeneous burn severity create a challenging classification scenario. The model attains an IoU greater than 0.7 in a mere 47\% of sites, signifying that while pixel-level classification demonstrates success, accurate spatial extent estimation remains challenging, particularly for complex fire perimeters exhibiting fingering patterns and unburned islands within the fire matrix.

\begin{table}[h!]
\centering
\caption{Performance metrics on European EMSR test sites}
\label{tab:results}
\begin{tabular}{lcccccc}
\toprule
\textbf{Site} & \textbf{Accuracy} & \textbf{Precision} & \textbf{Recall} & \textbf{F1} & \textbf{IoU} & \textbf{Dice} \\
\midrule
EMSR633 & 0.948 & 0.928 & 0.887 & 0.907 & 0.830 & 0.907 \\
EMSR749 & 0.974 & 0.871 & 0.894 & 0.883 & 0.790 & 0.883 \\
EMSR748 & 0.916 & 0.792 & 0.922 & 0.852 & 0.742 & 0.852 \\
EMSR747 & 0.797 & 0.353 & 0.974 & 0.519 & 0.350 & 0.519 \\
EMSR746 & 0.949 & 0.892 & 0.955 & 0.922 & 0.856 & 0.922 \\
EMSR745 & 0.959 & 0.963 & 0.541 & 0.693 & 0.530 & 0.693 \\
EMSR744 & 0.988 & 0.944 & 0.921 & 0.932 & 0.873 & 0.932 \\
EMSR740 & 0.982 & 0.858 & 0.828 & 0.843 & 0.728 & 0.843 \\
EMSR738 & 0.932 & 0.917 & 0.782 & 0.844 & 0.731 & 0.844 \\
EMSR737 & 0.990 & 0.900 & 0.812 & 0.854 & 0.745 & 0.854 \\
EMSR736 & 0.912 & 0.898 & 0.372 & 0.526 & 0.357 & 0.526 \\
EMSR735 & 0.989 & 0.784 & 0.880 & 0.829 & 0.708 & 0.829 \\
EMSR733 & 0.977 & 0.456 & 0.940 & 0.614 & 0.443 & 0.614 \\
EMSR731 & 0.997 & 0.953 & 0.417 & 0.580 & 0.409 & 0.580 \\
EMSR730 & 0.956 & 0.934 & 0.843 & 0.886 & 0.796 & 0.886 \\
EMSR729 & 0.993 & 0.333 & 0.496 & 0.399 & 0.249 & 0.399 \\
EMSR638 & 0.874 & 0.934 & 0.269 & 0.418 & 0.264 & 0.418 \\
\midrule
\textbf{Mean (N=17)} & 0.949 & 0.806 & 0.749 & 0.735 & 0.612 & 0.735 \\
\textbf{Std} & 0.052 & 0.211 & 0.231 & 0.186 & 0.220 & 0.186 \\
\bottomrule
\end{tabular}
\end{table}

\section{Conclusion}
This study demonstrates the remarkable potential of combining the AlphaEarth dataset with a Siamese U-Net architecture for global burned area mapping. The outcomes are particularly encouraging, given that the model, which was trained exclusively on MTBS data from the contiguous United States, achieved a mean accuracy of 94.9 percent when evaluated across 17 diverse European EMSR sites. This cross-continental generalization capability suggests that the learned fire signatures transcend regional boundaries, capturing fundamental spectral and textural characteristics of burned vegetation that remain consistent across different ecosystems. The AlphaEarth dataset proved instrumental in this success. The instrument's 64 spectral bands, which span optical and thermal infrared wavelengths, provide a comprehensive and nuanced representation of the electromagnetic spectrum. This multifaceted information enables the discernment of burned areas and spectrally similar features, facilitating a more precise and detailed analysis of the affected regions. A critical aspect of the dataset's preprocessing involves the elimination of common confounding factors that typically compromise the accuracy of operational burned area mapping systems. These confounding factors include cloud contamination, atmospheric effects, and sensor noise. This feature enabled us to prioritize model architecture and training strategies over the development of extensive data preprocessing pipelines. However, an in-depth analysis of the available data reveals persistent challenges that warrant further discussion. The issue of false alarms persists as a significant problem, particularly in agricultural regions where the cyclical harvesting of crops and the preparation of fields result in the generation of spectral signatures that bear a striking resemblance to those of fire damage. This issue is exemplified by the precision scores of 0.353 and 0.333, respectively, obtained by sites like EMSR747 and EMSR729. Furthermore, areas demonstrating accelerated post-fire vegetation recovery present detection challenges, as the spectral evidence of burning rapidly diminishes in regions exhibiting favorable growing conditions. The model demonstrates challenges with fragmented burn patterns and gradual severity transitions, where the binary classification framework may impose inherent limitations.
Subsequent research endeavors should explore several promising avenues. First, the implementation of a multi-class severity classification scheme, as opposed to binary detection, would facilitate a more comprehensive capture of the intricacies associated with fire impacts. In conclusion, the expansion of the training dataset to encompass fires from multiple continents would likely enhance the model's global applicability. The incorporation of explainable AI techniques has the potential to elucidate the spectral bands that contribute most significantly to detection accuracy, thereby facilitating more efficient operational implementations with reduced computational requirements.

\section*{Acknowledgments}

We thank the AlphaEarth Foundations team for providing the embedding dataset, 
which is available at \url{https://developers.google.com/earth-engine/datasets/catalog/GOOGLE_SATELLITE_EMBEDDING_V1_ANNUAL}. The MTBS program deserves recognition for maintaining the comprehensive burned area database that served as our training foundation (\url{https://www.mtbs.gov/}).  We also acknowledge the Copernicus Emergency Management Service Rapid Mapping program 
for providing invaluable validation data through their EMSR products 
(\url{https://emergency.copernicus.eu/mapping/list-of-components/emsr}). \
We acknowledge OpenAI’s ChatGPT-5 for assistance in improving the clarity and readability of the manuscript text.

\section{Supplemental Materials}
This section presents comprehensive visual results of burned area mapping for the remaining 14 European EMSR test sites not featured in the main manuscript. These results provide deeper insights into the model's performance variability across diverse geographic regions, fire regimes, and landscape characteristics.
As illustrated in Figure~\ref{fig:burned_area_visualization}, the comprehensive array of predictions across all test sites unveils discernible performance patterns that are associated with the unique characteristics of regional fires. The confusion matrix overlays facilitate pixel-level error analysis, where true positives (white) denote correctly identified burned areas, true negatives (black) indicate properly classified unburned regions, false positives (red) highlight commission errors, and false negatives (blue) denote omission errors.\

Notable observations from the extended results include:

\textbf{High-performance sites:} High-performance sites The performance of the EMSR633 and EMSR638 models is characterized by their ability to delineate boundaries with minimal edge artifacts. This is evident in the compact true positive regions and negligible false negative halos observed. These sites are distinguished by the presence of vegetation types that are genetically similar and by the presence of clear burn severity gradients, which facilitate the establishment of robust discrimination methods.

\textbf{Challenging detection scenarios:} EMSR729 displays severe commission errors, indicated by substantial red regions, despite high pixel accuracy. This suggests the potential for spectral confusion with agricultural activities or seasonal vegetation senescence. EMSR731, conversely, demonstrates an alternative pattern characterized by conservative detection accompanied by substantial omission errors (i.e., blue regions), particularly evident in low-severity burn zones.

\textbf{Fragmentation effects:} As evidenced by the analysis of sites EMSR745 and EMSR747, the model exhibits a clear challenge in managing discontinuous burn patterns. The presence of false negative pixels, distributed haphazardly within the designated burn perimeters, signifies the challenges associated with detecting vegetation that has been partially burned or has undergone rapid post-fire recovery. EMSR747's high rate of false positives, primarily evident in the red prediction map, signifies a complete failure of the model in complex terrain.

\textbf{Edge precision analysis:} The specimens designated EMSR736, EMSR737, and EMSR740 exhibited superior edge accuracy, characterized by distinct transitions between the burned and unburned regions. The minimal false positive/negative pixels along fire boundaries suggest effective learning of edge features through the Siamese architecture's temporal comparison mechanism.

\textbf{Scale-dependent performance:} Large contiguous burns (EMSR748, EMSR749) demonstrate balanced precision-recall with minimal salt-and-pepper noise, while smaller fires (EMSR730, EMSR733) exhibit augmented relative error proportions due to the fixed spatial resolution constraining sub-pixel burn detection.

\begin{figure}[p]
\centering
\makebox[\textwidth][c]{\includegraphics[width=1.1\textwidth]{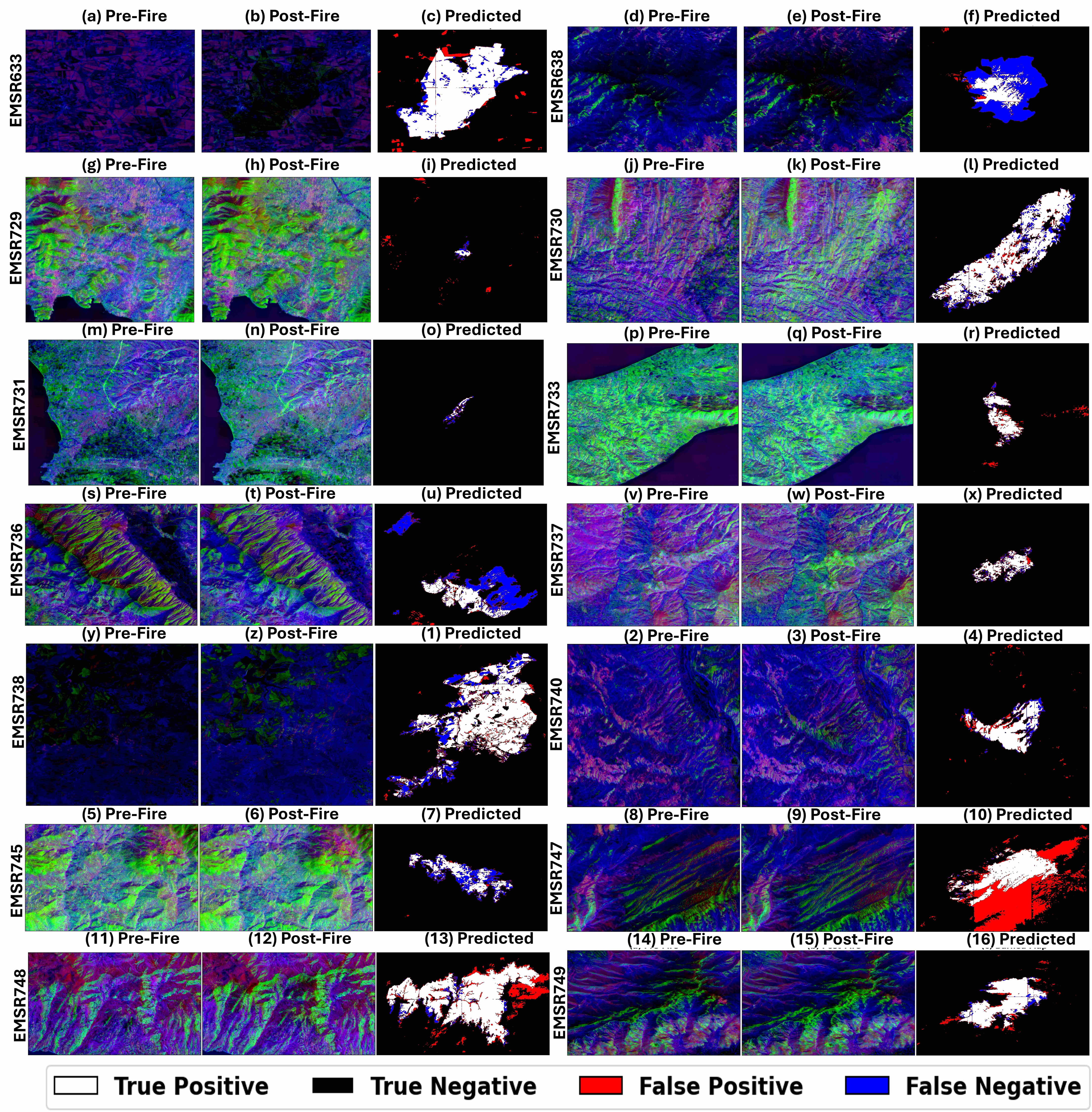}}
\caption{Comprehensive visual validation results for 14 European EMSR test sites showing burned area detection performance. Each row presents a different test site with columns showing: (a,d,g,j,m,p,s,v,y,2,5,8,11,14) pre-fire false-color composite imagery, (b,e,h,k,n,q,t,w,z,3,6,9,12,15) post-fire imagery revealing burn scars, and (c,f,i,l,o,r,u,x,1,4,7,10,13,16) predicted burned area maps with confusion matrix overlay. }
\label{fig:burned_area_visualization}
\end{figure}

\bibliographystyle{ieeetr} 
\bibliography{references}   

\end{document}